# EXTENSION OF HIDDEN MARKOV MODEL FOR RECOGNIZING LARGE VOCABULARY OF SIGN LANGUAGE

Maher Jebali[1], Patrice Dalle[2], and Mohamed Jemni[1]

[1]Research Lab. LaTICE – ESSTT University of Tunis – Tunisia

maher.jbeli@gmail.com

[2]Research Lab. IRIT Univ. of Toulouse3 – France

patrice.dalle@irit.fr

*ABSTRACT*

*Computers still have a long way to go before they can interact with users in a truly natural fashion. From a user's perspective, the most natural way to interact with a computer would be through a speech and gesture interface. Although speech recognition has made significant advances in the past ten years, gesture recognition has been lagging behind. Sign Languages (SL) are the most accomplished forms of gestural communication. Therefore, their automatic analysis is a real challenge, which is interestingly implied to their lexical and syntactic organization levels. Statements dealing with sign language occupy a significant interest in the Automatic Natural Language Processing (ANLP) domain. In this work, we are dealing with sign language recognition, in particular of French Sign Language (FSL). FSL has its own specificities, such as the simultaneity of several parameters, the important role of the facial expression or movement and the use of space for the proper utterance organization. Unlike speech recognition, Frensh sign language (FSL) events occur both sequentially and simultaneously. Thus, the computational processing of FSL is too complex than the spoken languages. We present a novel approach based on HMM to reduce the recognition complexity.*

## 1. INTRODUCTION

The focus of Human-Computer Interaction (HCI) is becoming more significant in our life. With the progress of computer science, the existing devices used on HCI (keyboard, mouse, ) are not satisfying our needs nowadays. Many designers attempt to make HCI more natural and more easier. They introduced the technique of Human-to-Human Interaction (HHI) into the domain of HCI, to reach this purpose. In this context, using hand gestures is among the richest HHI domain since everyone uses mainly hand gestures to explain ideas when he communicates with others. Communication with hand gestures becomes very important and much clearer with the consideration of sign language. Sign language (SL) is the natural way of communication among deaf persons. As letters and words in natural languages, we find the corresponding elements on sign languages, which are the movements, gestures, postures and facial expressions. Many researchers exploited hand gestures in other application fields such as hand tracking [2] and interactive computer graphics [3]. Similarly, several studies have focused automatic sign language recognition [1] [2] so as to facilitate communication between hearing and deaf persons and to improve Human-Computer Interaction systems. Several works in the domain of automatic sign language recognition have been interested in the manual information such as the trajectories of both hands. In this context, we are presenting in this paper, our approach for modelling the manual information. In forthcoming papers, we will present others components of our sign language recognition system, such as non-manual signs (head movement, gaze…) and especially the facial expression that gives additional information to convey the true meaning of sign [24]. This article is structured as follows. Section 2 gives an overview of related works in this study. Section 3 gives an overview of sign recognition. Section 4 presents the main problems of automatic sign language recognition. Section 5 details the novel extension of HMM that we proposed for reducing the complexity task.

## 2. RELATED WORKS

In order to recognize human gestures by motion information [4] [28], many methods, based on trajectories and motions, have been proposed to analyze these gestures. In [5], Bobick and Wilson adopted a state-based method for gestures recognizing. They used a number of samples, for each gesture, to calculate its main curve. Each gesture sample point was mapped to the length of arc at curve. Afterword, they approximated the discritized curve using a uniform length in segments. They grouped the line segments into clusters, and to match the previously learning state of sequences and the current input state, they used the algorithm of Dynamic Time Warping. A 3D vision-based system was described in [6] for American Sign Language (ASL) recognition. They used a method based on 3D physics tracking to prepare their data samples, which were used to train HMMs. They used the extended Kalman filter to predicate the motion of the model. This input of the HMM method was a set of translation and rotation parameters. Subsequently, Vogler and Metaxas proposed an extended HMM called Parallel Hidden Markov Model to overcome the problem of large-scale vocabulary size. In [23], they took into account the sentence spatial structure. They have presented a general sign language model of sentences that uses the composition of the space of signing as a representation of both the meaning and the realisation of the utterance. Fels and Hinton [7] used in their recognition system a data gloves as an acquisition tool, and they have chosen 5 neural networks to classify 203 signs. For continuous ASL recognition, Haynes and Jain used a view-based approach [9]. [16] Starner and al. extracted 2D features from a sign camera. They obtained 92%. In [20], the authors used many characteristics to detect the borders of Japanese SL words, like orientation, hand shape and position. [26] [27] have developed a language recognition application based on 3D continuous sign. They used the Longest Common Subsequence method for sign language recognition instead of HMM, which is a costly double stochastic process. A method of dynamic programming was used by the authors of [11] [12] to recognize continuous CSL. They used a data gloves as acquisition devices and Hidden Markov Model as a method of recognition. Their system recognized 94.8%. Previous research on automatic recognition of sign language mainly focused on the domain of the dependent signer. A little work has been devoted to independent signer. Indeed, the sign language recognition systems based on independent signer have a promising trend in practical systems that can recognize a different signers SL. But the problem of the recognition of signer independent is hard to resolve regarding the great obstacle from different sign variations.

Table 1. Gesture recognition methods

| Method | Capture Methods | Recognition algorithm | Summary |
|---|---|---|---|
| [14] | Image moments | HMM | Recognize a set of 6 users defined hand gestures in a restrictive back-ground |
| [16] | Colored glove or skin tone blobs | HMM | Specifically designed to recognize a set of 40 ASL |
| [8] | Data glove | Multilayer neural network network | Extract motion information using data gloves and train neural networks for speech synthesizer control |
| [15] | Colored marker | HMM and predicate calculus | Recognize 6 visual events based on predicate calculus in which each predicate is maximally estimated from trained hidden Markov models |
| [5] | Data glove template | Principal curve and DTW | Recognize various gestures extracted from using motion trajectories magnetic sensors or Eigen templates of hand images in restrictive back-ground |
| [13] | Data glove | Parametrized HMM | Extract 3D locations of hand using stereo cameras and skin color to recognize a set of 32 size gestures |
| [1] | 3D camera system with physics model | HMM | Develop parallel HMMs to recognize 22 ASL gestures with a 3D camera system |

## 3. SIGN lANGUAGE OVERVIEW

### 3.1 Lexical Meaning Expression

The linguists of sign language make out the basic components of a sign as corresponding of the handshape, location, hand orientation and movement. Handshape means the fingers configurations, location means where the hand is localized relatively to a parts of body and orientation means the direction in which the fingers and palm are pointing. The movement of hand draws out a trajectory in the signing space. Stokoe, who proposed the first phonological sign language model, accentuates the simultaneous presentation of these components. The model of Liddell and Johnson [18] accentuated the sequentiality organization of these components. They defined the segments of movement as short periods during which some sign component is in transition. The segments of hold are short periods when all these components are static. The aim of many novel models is to represent the sequential and the simultaneous signs structure and it would look that the adopted SL recognition system must be capable to model sequential and simultaneous structures. Different researchers used constituent parts of signs for classification systems. All signs parts are very important as manifested by the signs unit existence, which differs in only one of the principal components. When signs take place in a continuous video to product utterances, the hand must change the position from the ending of one sign to the starting of the next sign. At the same time, the orientation and the handshape also change from the ending orientation and handshape of one sign to the starting orientation and handshape of the next sign. These periods of transition from one position to another are called movement epenthesis and they do not represent any part of signs. The production of continuous signs processes with the same effects of the co-articulation in speech language, where the preceding and succeeding signs utter the sign appearance. However, these movements epenthesis are not presented in all signs. Hence, the movement epenthesis presents often times during the production of a continuous signs and should be analyzed first, before dealing with the different phonological aspects. Extraction and classification feature methods are affected by some signing aspects, especially for approaches based on vision. First, when a sign gesture is produced, the hand may be placed in different orientations with consideration to the morphological signers body and a fixed orientation of the hand cannot be supposed. Second, different movements types are included in signs production. Generally, the movement represent the whole hand drawing a general 3D trajectory. However, there are many signs whose only represent local movements, such as changing the orientation of the hand by moving the fingers or deforming the wrist. Third, the two hands often occlude or touch each other when captured from only one field of view and the hands partially occlude the face in some signs. Hence, handling the occlusion is an important step.

## 3.2 Grammatical Processes

The different changes to the sign aspect during the production of continuous signs do not change the sign meaning. However, there are other amendments to one or more sign components, which affect the meaning of sign, and these are concisely described in this section. Usually, the meanings transmitted through these changes are related to the verbs aspects that include recurrence, frequency, permanence, intensity and duration. Furthermore, the movement of sign can be changed through its tension, rhythm, trajectory shape and rate. [17] list between 8 and 11 types of possible inflections for temporal aspect. The agreement of person (first person, second person, or third person) is an another type of inflection that can be presented. Here, the verb designates its object and subject by a modification in the direction of movement with corresponding modifications in its start and end position, and the orientation of hand. Verbs can be changed simultaneously for number agreement and person. Many other samples of grammatical aspects which turn out in orderly changes in the appearance of sign involve categorical derivation of nouns from verbs, inflections, compound signs and numerical incorporation. Categorical inflections are used for the accentuation intention and are shown across repetition in the movement of sign, with tension throughout. In a classic system of sign language recognition, data are dealt together in only one block. However, both hands would be represented as a single block. At this level, us show that this system type of recognition looks not well adapted to features of SL. To facilate sign language analysis sentences and to associate lexical level to superior level, the authors of [25] have proposed a computational model for signing space construction that can be related to the grammar of sign language.

## 3.3 Non-manual gestures

For non-manual, we mean the movement of the head, eyebrows, eyelids, cheeks, nose, mouth, shoulders and torso and the direction of gaze. This section presents the non-manual gestures based on the current knowledge, particularly their linguistic functions, to propose a reflection at all levels of sign languages. The non-manual elements, in particular, the facial expression have long been put aside by linguists. Subsequently, the non-manual elements evoked on many occasions for their linguistics roles but without having been specific studies. The non-manual elements intervene at the lexical level in four cases :
- In the constitution of a standard sign.

- The distinction between two signs.
- The distinction between the object and its action and/or use.
- In the modifiers realization (adjectives, adverbs, etc.).

The non-manual elements intervene also on the utterances. It has a comparable role to the punctuation in writing or the tone of the speech. Thus, it is essential to the proper understanding of the final message. These elements present also an important role in the segmentation of utterances and in particular to delimit syntactic groups.

The figure fig.1 shows two signs which have the same manual elements and distinguished through the noon manual elements.

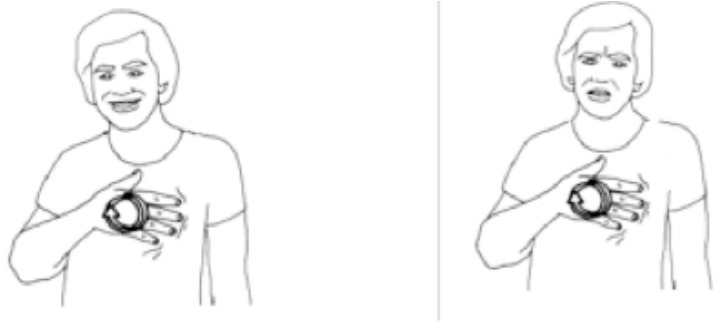

Figure 1: (content) and (pain in the heart)

## 4. MANUAL SIGN RECOGNITION ISSUES

### 4.1 Lexicon Complexity

Complexity is a relevant issue in sign language recognition systems when the lexicon has a large size. Since a sign consists of numerous entities occurring together, the lexicon of SL can incorporate a large number of signs [19]. We can obtain HS possible handshapes for each hand, M different movements, L locations, and O orientations. Thus HS*L*O*M possible signs. Furthermore, both hands can be associated together, thus we can have a very large lexicon. Modeling a sign language recognition system for such a lexicon is challenging task. In fact, the signers can not represent all possible combinations. Signs parameters are associated following semantic and grammatical rules. Nevertheless, the resulting lexicon size remains too large. To figure out this problem, instead processing signs parameters like only one block, we decide to deal them separately as in [20] [21]. We can process like in speech recognition system (SRS). In SRS, the phonemes are recognized instead whole words. But in speech language, phonemes are always successive, while in the parameters of SL, they are often simultaneous. Therefore, establishing a recognition system remains arduous.

### 4.2 Role of both hands in signing space

There are three kinds of interactions between right and left hands. Either both hands creates together a sign, or only one hand produce a sign, or each hand creates a sign and those signs can be completely independent or have a relationships. The problem is being able to set aside each instance. Thus, we will not recognize a sign produced with both hands when there are two different signs executed at the same time. To differentiate these two cases, the similarity parameters measure between both hands is unusable, because Signs produced with two hands, may be entirely synchronous, as well as correlated and not having a similarity between parameter. It is necessary to find the two other existing relationships between hands. When both hands produces two different signs with a relationship of dominant / dominated hand, there are points of synchronization between the different signs. However, both hands will be structured in a manner that supply informations. Nevertheless, during a brief time, the two hands are structured in order to supply the spatial informations. Thus, we demand to be capable to capture this especial period in order to represent this spatial relationships. Even if the information transmitted by the two hands is only spatial, the relationship between both hands is spatio-temporal.

### 4.3 Translation of non-standard Sign

The third issue concerns the non standard signs translation, and this is a very difficult problem. Currently, only a little part of those signs has been accounted by systems dealing with sign recognition [22] [23]. Classifiers, size and shape specifiers represent a part of non-standard signs. The utterance and its context give a meaning. Classifiers are frequently associated to a standard sign which has been previously produced, so it will be an easy task to give a sense to the sign. Unluckily, some classifiers are linked to signs which are produced after them or they are linked to descriptions made with specifiers. Yet, we can develop a possible set of classifiers for signs. Detecting a sens for specifier is more difficult, because it is particular to what it describes. Therefore, from one characterization to another, signs can widely vary. Unlike the limited lexicons in oral languages, SL contains many signs which can be invented in accordance with needs. The signs specifiers parameters are linked to what they describe. Therefore, new signs can appear for each description. Moreover, the way which the description is done depends on the signers cultural and social context. We need to analyze the utterance and its context to understand these non standard signs. Also, we need to analyze different parameters of those signs for associating them with what is being described. This is not evident, and can only be dealt at the semantic and syntactic level. Primarily, sign language recognition systems work at the lexical level, thus, they are not capable to deal with such signs which are declined or badly represented. We need to use a higher level analyzer to deal these stored signs.

## 5. HIDDEN MARKOV MODEL

From a set of N states $C_i$, the transitions from two states can be described, at each time step t, as a stochastic process. The probability of transition to reach state Ci in the first time step is mentioned as the probability of transition $a_{ij}$ of state $C_i$ to another state $C_j$ only depends on the preceding states, this process is called Markov chain. The additional hypothesis, that the current transition only depends on the preceding state leads to a first order Markov chain. A second stochastic process can be defined, that produces a symbol vectors x at each time step t. The Using HMM for sign recognition is motivated by the successful application of probability of emission of a vector x wich not depends on the way that the state was attained but only depends on the current state. The density of emission probability $y_i(x)$ for vector x at state Ci can either be continuous or discrete. This process is a doubly stochastic and called a Hidden Markov Model (HMM) if the sequence of state are not observable, but not the vectors x. A HMM H is defined as follows H = (x, A, B). x represent the vector of the initial probabilities of transition , the NxN matrix A denotes the probabilities of transition a, from state $C_i$ to $C_j$ and finally, B represents the emission densities vector yi(x) of each state.

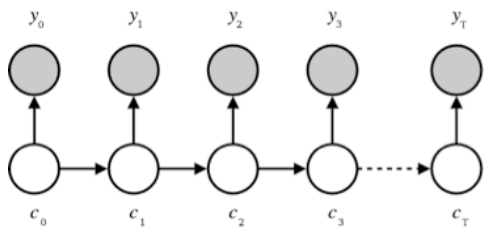

**Figure 2: Hidden Markov Model**

Using HMM for sign recognition is motivated by the successful application of the techniques of Hidden Markov Model to speech recognition issues. The similar points between speech and sign suggest that effective techniques for one problem may be effective for the other as well. First, like spoken languages, gestures vary according to position, social factors, and time. Second, the movements of body, like the sounds in speech, transmit certain meanings. Third, signs regularities performances while speaking are similar to syntactic rules. Therefore, the methods elaborate by linguistic may be used in sign recognition. Sign recognition has its own characteristics and issues. Meaningful signs may be complex to deal, containing simultaneous movements. However, the complicated signs should be facilely specifiable. Generally, signs can be specified either by description or by example. Earlier, each system has a training stage in which samples of different signs are collected to train the models. These models are the all gestures representations that will be recognized by the system. In the latter specification method, a description of each sign is written with a a formal language of SL, in which the syntax is specified for each sign. Evidently, the description method has less flexibility than the example method. One potential inconvenience of example specification is difficult to specify the eligible variation between signs of a given class. If the parameters of the model were determined by the most likely performance criterion, this problem would be avoided. Because sign is an expressive motion, it is evident to describe a movement across a sequential model. Based on these criterions, Hiddem Markov Model is appropriate for sign recognition. A multi-dimensional HMM is capable to deal with multichannel signs which are general cases of sign language recognition.

## 5.1 Independent channels

In speech, phonemes occur in sequence, but in sign languages can occur in parallel. Some signs are produced with two hands, so they must be modelled. Many approaches highlighted the massive complexity of the simultaneous aspects modelling of sign languages. It is infeasible to model all these combinations for two reasons. First, the computational and the modelling viewpoint. Second, it would be impossible to gather sufficient data for all these combinations. The main solution is to find a way to extinguish the simultaneous events from one another. However, it is viable to seem at only one event at a time. The idea of this approach comes from many researchers in linguistic field into sign languages, especially into French Sign Language (FSL), most of which concentrated on a particular various aspect of the features. Each hand had their own features such as the hand shape, orientation or position. Splitting the manual and non-manual components within independent channels allows a major reduction in the modelling complexity task. Thus, it is not needful to envisage all possible phonemes combinations. Instead, we model the phonemes in a single channel. Each channel contain a reduced number of different Hidden Markov Models to represent the phonemes. Therefore, phonemes combinations are easy to reliate together at time of recognition, especially in association with the parallel HMM.

## 5.2 Three parallel channels

Conventional HMMs are an unsuitable choice for SL modelling for many reasons. First, they are only able to model single event. However, they should merging the different channels into a single event, forcing them to acquire in a coupled manner, which is poor for many systems that demand modelling of different simultaneous processes. For example, if a sign, produced with only one hand, precedes a sign performed with both hands, the dominate hand moves often to the position requisite by the sign performed with both hands before the dominant hand starts to make it, at a slightly inconclusive point in time. If the different channels wee too closely coupled, the dominate hand movement would be impossible to capture. Unfortunately, this extension cannot solve the key modelling complexity problem of French sign language. This approach needs to train the different interactions between these channels. Thus, there would require to be sufficient training examples usable for all possible interaction between FSL phonemes through channels. Therefore, it is necessary to make a novel extension to HMMs that allow phonemes combinations decoupling, and to model them at recognition moment, instead of training moment. We describe now the proposed three channels parallel HMMs as a possible solution. We model each channel $C_i$ with independent HMM with separate output.

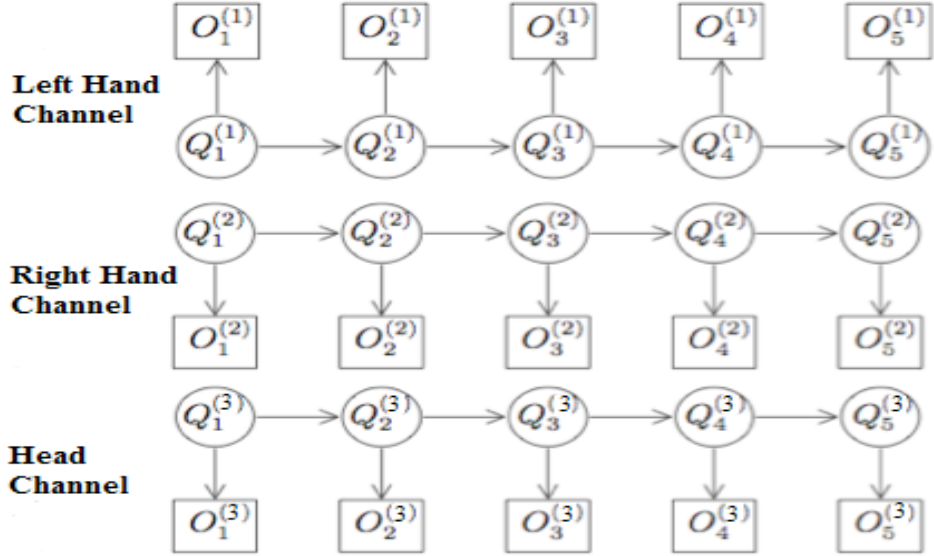

Figure 3: Three channels parallel HMM

The separate channels of parallel HMM progress independently from one to another and they have autonomous output. However, it is possible to train independently the different channels, and to set the different channels together at recognition moment. Using this novel approach, we have to reduce the FSL modelling complexity, because, instead to perform all phonemes possible combinations, we are invited to consider each phoneme by itself. Thus, the total number of hidden Markov model needed is only the sum, instead of the product of all phonemes. At recognition time, it is indispensable to consolidate some informations from the HMM representing the three different channels. The recognition algorithm should find the maximum joint probability of the three channels that is :

$$\max_{Q^{(1)},...,Q^{(C)}} \left\{ \log P(Q^{(1)},...,Q^{(C)}, O^{(1)},...,O^{(C)} | \lambda^{(1)},...,\lambda^{(C)}) \right\}$$

$$\max_{\mathbf{Q}^{(1)},...,\mathbf{Q}^{(C)}} \left\{ \log P(\mathbf{Q}^{(1)},...,\mathbf{Q}^{(C)}, \mathbf{O}^{(1)},...,\mathbf{O}^{(C)} | \lambda^{(1)},...,\lambda^{(C)}) \right\} = \max_{\mathbf{Q}^{(1)},...,\mathbf{Q}^{(C)}} \left\{ \sum_{c=1}^{C} \log P(\mathbf{Q}^{(c)}, \mathbf{O}^{(c)} | \lambda^{(c)}) \right\}.$$

Where $Q^{(c)}$ is the state sequence of channel c, $1 \leq c \leq C$, with observation sequence $O^{(c)}$ through the HMM network $\lambda^{(c)}$. Recall that this observation sequence corresponds to some unknown sign sequence in channel c, which is to be recognized. Because in Parallel HMMs the channels are independent, the merged information consists of the product of the probabilities of the individual channels or the sum of the log probabilities.

## 6. CONCLUSION

Seeing that the two hands are represented by independent channels of communication produces different relationships between signs produced by left hand and those produced by right hand and the important role of the head to complete the sign meaning. We can not neglect those relationships because they convey in formations which are very needful to the recognition system. We have decribed a new method for FSL recognition, wich is based on phonological modeling of sign language. We have presented Parallel Hidden Markov Model with three independent channels wich represent respectively the right hand, the left hand and the head. This recognition method has the potential to manage a large-scale vocabularies than the majority of current and past research that have been used into sign language recognition.